\def\year{2021}\relax
\documentclass[letterpaper]{article} % DO NOT CHANGE THIS
\usepackage{aaai21}  % DO NOT CHANGE THIS
\usepackage{times}  % DO NOT CHANGE THIS
\usepackage{multirow}
\usepackage{multicol}
\usepackage{array}
\usepackage[switch]{lineno}
\usepackage[utf8]{inputenc}
\usepackage{helvet} % DO NOT CHANGE THIS
\usepackage{courier}  % DO NOT CHANGE THIS
\usepackage[hyphens]{url}  % DO NOT CHANGE THIS
\usepackage{graphicx} % DO NOT CHANGE THIS
\usepackage{amsmath,amssymb,amsfonts}
\usepackage{tabularx}
\usepackage{booktabs}
\usepackage{natbib}  % DO NOT CHANGE THIS AND DO NOT ADD ANY OPTIONS TO IT
\usepackage{caption} % DO NOT CHANGE THIS AND DO NOT ADD ANY OPTIONS TO IT
\title{Two-Stream Appearance Transfer Network for Person Image Generation}
\author{Chengkang Shen \textsuperscript{1 $\ast$}, {Peiyan Wang} \textsuperscript{2} \thanks{The authors have contributed equally to this work.},Wei Tang \textsuperscript{1}}
\affiliations{
    \textsuperscript{\rm 1}University of Illinois at Chicago
    
    \textsuperscript{\rm 2}Purdue University\\
    \{cshen26,tangw\}@uic.edu, wang5035@purdue.edu

}
\begin{document}
%\linenumbers  %
\maketitle
\begin{abstract}
Pose guided person image generation means to generate a photo-realistic person image conditioned on an input person image and a desired pose. This task requires spatial manipulation of the source image according to the target pose. However, the generative adversarial networks (GANs) widely used for image generation and translation rely on spatially local and translation equivariant operators, i.e., convolution, pooling and unpooling, which cannot handle large image deformation. This paper introduces a novel two-stream appearance transfer network (2s-ATN) to address this challenge. It is a multi-stage architecture consisting of a source stream and a target stream. Each stage features an appearance transfer module and several two-stream feature fusion modules. The former finds the dense correspondence between the two-stream feature maps and then transfers the appearance information from the source stream to the target stream. The latter exchange local information between the two streams and supplement the non-local appearance transfer. Both quantitative and qualitative results indicate the proposed 2s-ATN can effectively handle large spatial deformation and occlusion while retaining the appearance details. It outperforms prior states of the art on two widely used benchmarks.
\end{abstract}
\section{Introduction}
Pose guided person image generation aims to transform a person image from a source pose to a target pose while retaining the appearance details. It serves as a fundamental tool for several practical applications such as image editing, video generation and data augmentation for person re-identification and action recognition \cite{yang2018pose,zhu2019progressive,qian2018pose}. This task is very challenging especially in case of large pose transform, occlusion and complex texture.

\begin{figure}[t]
\begin{center}
\includegraphics[width=0.47\textwidth]{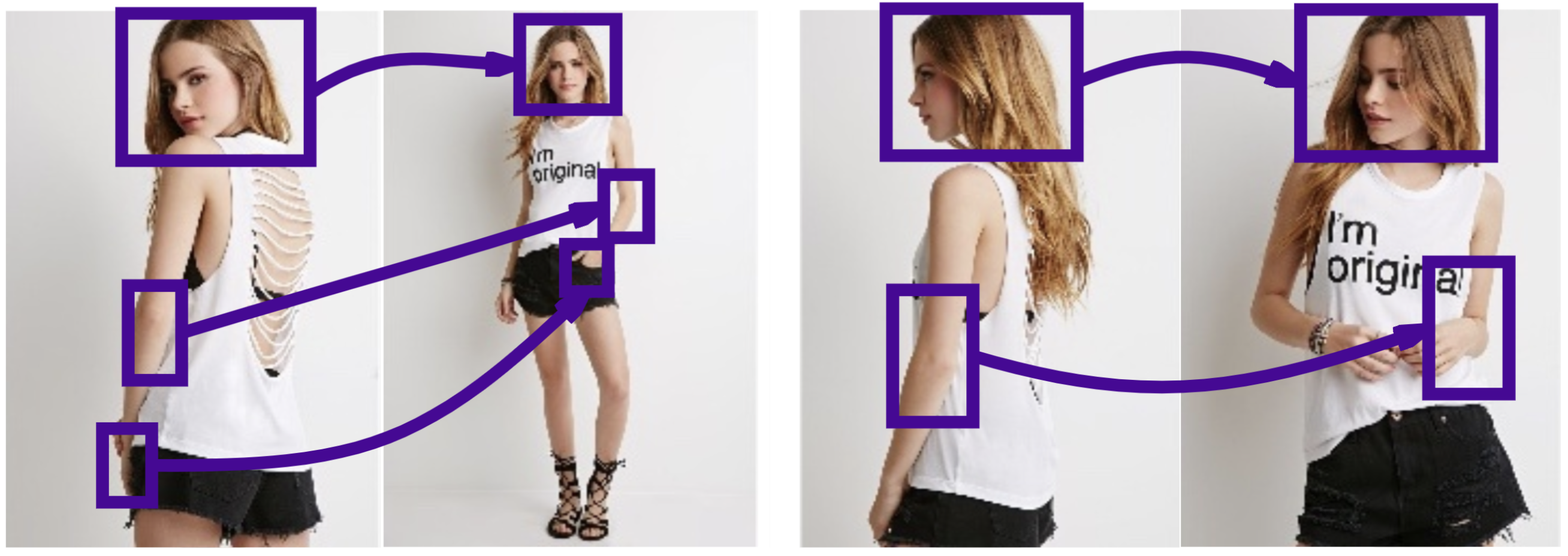}
\end{center}
\caption{Transferring the pose of a person requires spatial manipulation of the source image.}
\label{motivation}
\end{figure}

Convolutional neural networks (CNNs) \cite{lecun1998gradient} and their variants \cite{,ronneberger2015u,he2016deep}, trained in an adversarial fashion \cite{goodfellow2014generative}, have been widely used for image generation and translation \cite{isola2017image,ledig2017photo,johnson2016perceptual,zhu2017unpaired,mirza2014conditional}. However, since CNNs are composed of spatially local and translation equivariant operators, i.e., convolution, pooling and unpooling, they do not have an explicit mechanism to handle articulated body deformation, as illustrated in Fig. \ref{motivation}.
To resolve this difficult issue, two strategies have been adopted in  prior person image generation approaches, i.e., parametric geometric transformation and nonparametric dense flow. For example, Siarohin et al. \cite{siarohin2018deformable} apply an affine transformation to the features of each body part region to deal with pixel-to-pixel misalignment caused by the pose difference. However, it cannot handle occlusion or out-of-plane rotation well. Some methods \cite{han2019clothflow,liu2019liquid} predict the dense flow field between the source and target images and apply it to warp the feature maps. However, since the flow is predicted via a CNN, it cannot account for large or non-local motion.

\begin{figure*}[t]
\begin{center}
\includegraphics[width=1\textwidth]{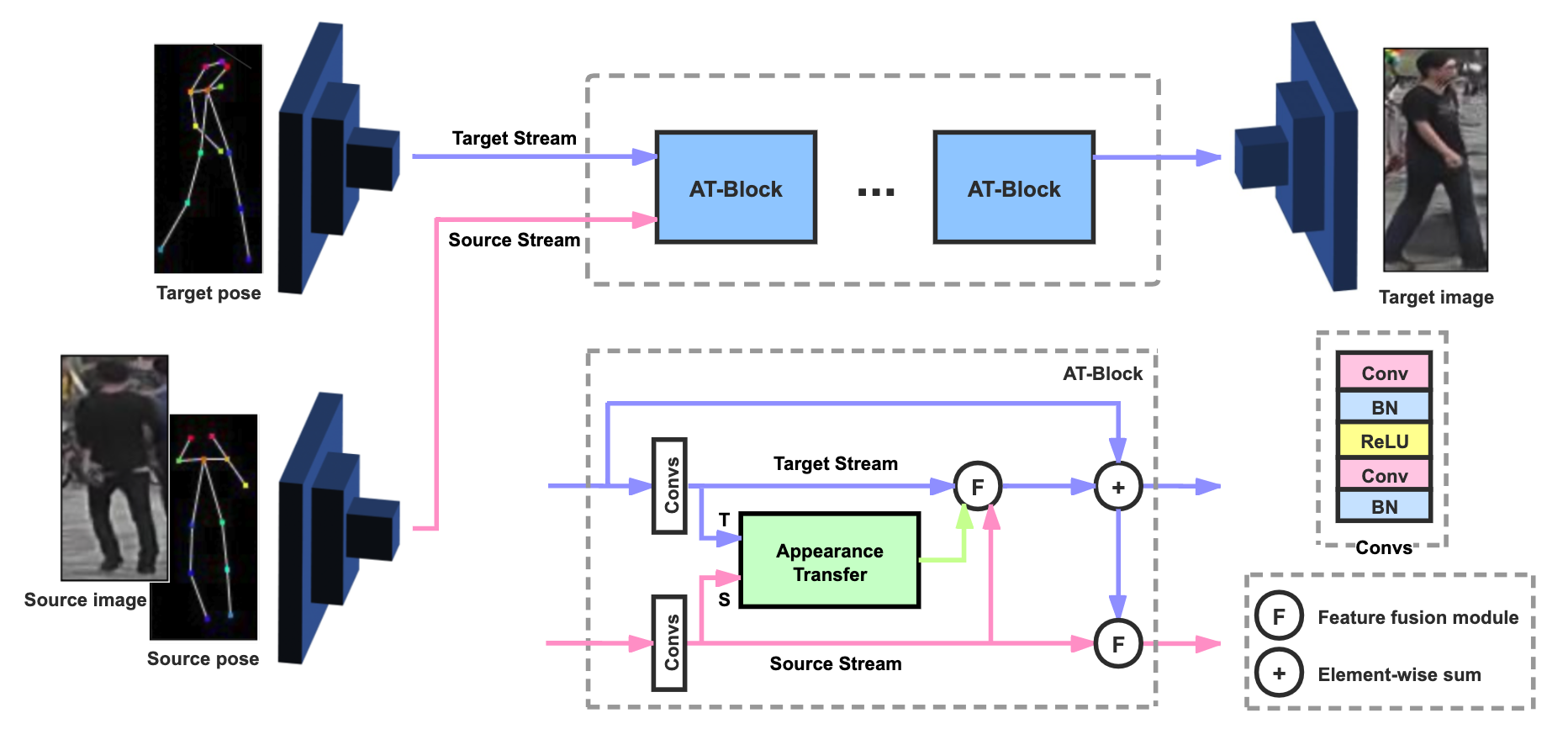}
\end{center}
\caption{An overview of the proposed two-stream appearance transfer network (2s-ATN). It is a multi-stage architecture consisting of two streams. The target and source streams respectively take as input the target pose and the source pose and image, and pass them through convolutional encoders. Each stage is an appearance transfer block (AT-block). It consists of an appearance transfer module (detailed in Fig. \ref{AT-module}) and two two-stream feature fusion modules. We consider two fusion functions, i.e., concatenation and summation, and find the former works better for both fusion modules. The target feature map from the last AT-block passes through a convolutional decoder to generate a new person image with the same appearance as the source image but in the target pose.}
\label{generator}
\end{figure*}

This paper introduces a novel two-stream appearance transfer network (2s-ATN) for person image generation. It can effectively handle large pose transform and occlusion while retaining the appearance details. As illustrated in Fig. \ref{generator}, the 2s-ATN is a multi-stage architecture consisting of a source stream and a target stream. The two streams respectively take as input the source pose and image and the target pose. 
At each stage is an appearance transfer block (AT-block). It consists of a novel appearance transfer module (AT-module) and several two-stream feature fusion modules. 
The AT-module finds the dense correspondence between the two-stream feature maps and then transfers the appearance information from the source stream to the target stream. Unlike the parametric geometric transformation or the nonparametric flow field, our AT-module is inspired by the self-attention \cite{vaswani2017attention} and performs a \emph{query-and-transfer} procedure. But different with the self-attention, the queries, keys and values in our AT-module have explicit semantic meaning, and they are specially designed for pose-guided appearance transfer. Specifically, the feature vector at each spatial location in the target stream is taken as a \emph{query} to match the \emph{key} feature vectors in the source stream so that the corresponding appearance \emph{values} in the source stream can be transferred to the desired location in the target stream. Meanwhile, the two-stream feature fusion modules allow local information exchange between the two streams to supplement the non-local appearance transfer. Finally, the target stream outputs a generated image with the source appearance but in the target pose.

Ablation study indicates that the proposed 2s-ATN can effectively handle large spatial deformation and occlusion, and improve the quality of the generated person images.
Experimental results on two benchmark datasets, i.e., Market-1501~\cite{zheng2015scalable} and DeepFashion~\cite{liu2016deepfashion}, demonstrate the proposed approach outperforms state-of-the-art methods both qualitatively and quantitatively. 

The main technical contribution of this paper is the 2s-ATN and the AT-module. The 2s-ATN is a novel two-stream and multi-stage network architecture for person image generation. It progressively transfers the appearance from the source stream to the target stream guided by their spatial correspondence. Each stage combines a non-local AT-module and several local two-stream feature fusion modules. The proposed AT-module is the first of its kind to use the target stream to query and transfer the source stream. The local and non-local modules are complementary, and they together enable the network to effectively handle large pose deformation and occlusion.

%------------------------------------------------------------------------]

\section{Related Work}

%{\bf Person image Generation.} 
\subsection{Person Image Generation}
The task of pose guided person image generation was first introduced by~\cite{ma2017pose}. Their two-stage network first generates a coarse target image and then refines it in an adversarial way. 
\cite{ma2018disentangled} disentangles the foreground, background and pose information, and then manipulates them to get the desired pose. The controllability of the generation process is improved, but the quality of the generated image is reduced.
~\cite{esser2018variational} combines VAE~\cite{kingma2013auto} and U-Net~\cite{isola2017image} to distinguish the appearance and pose of a person image. However, it is difficult to represent the appearance features as a low-dimension underlying code, which unavoidably loses information.
~\cite{siarohin2018deformable} introduces deformable skip connections to transform the texture spatially. It uses a set of local affine transformations to decompose the overall articulated body deformation. However, it cannot handle occlusion or out-of-plane rotation well.
%~\cite{albahar2019guided} presents a bi-directional feature transformation (bFT) to to utilize the constraints of the semantic guidance better.
The pose attention transfer network (PATN)~\cite{zhu2019progressive} consists of an image stream and a pose stream, and it uses an attention mask to enhance the feature maps. However, it only processes features locally, and there is no explicit geometric manipulation or appearance transfer of the source image. By contrast, our 2s-ATN consists of a source stream and a target stream and explicitly finds their spatial correspondence to perform appearance transfer.

Most recently, 
~\cite{tang2019cycle} proposes a cycle-in-cycle GAN, which is a cross-modal framework exploring joint exploitation of the keypoint and the image data in an interactive manner.
~\cite{ren2020deep} introduces a differentiable global-flow local-attention framework to reassemble the inputs at the feature level.
~\cite{men2020controllable} proposes the attribute-decomposed GAN, which means to embed human attributes into the latent space as independent codes and thus achieve flexible and continuous control of attributes via mixing and interpolation operations in explicit style representations. 
~\cite{huang2020generating} introduces an appearance-aware pose stylizer, which generates human images by coupling the target pose with the conditioned person appearance progressively.
~\cite{lathuiliere2020attention} employs the local attention mechanism to select relevant information from multi-source human images for human image generation.
RATE-Net~\cite{yang2020region} leverages an additional texture enhancing module to extract appearance information from the source image and estimate a fine-grained residual texture map. This helps refine the coarse estimation from the pose transfer module.
~\cite{gao2020recapture} proposes a portrait photo recapture system with two modules that complement each other from both intra-part and inter-part perspectives to easily transform their portraits to the desired posture.

Several approaches adopt DensePose~\cite{alp2018densepose}, 3D pose ~\cite{li2019dense}, or human parsing~\cite{dong2018sof} to generate person images since they contain more information, e.g., the body part segmentation or depth. However, the keypoint based pose representation is much cheaper to obtain and more flexible. Therefore, we prefer to use a keypoint-based representation. 

\subsection{Self-attention}
The self-attention ~\cite{cheng2016long,parikh2016decomposable} was first introduced for natural language processing. It calculates the response of a certain position in the sequence by paying attention to all positions in the same sequence.
~\cite{vaswani2017attention} proves that the machine translation model could obtain the state-of-the-art results using the self-attention. 
~\cite{parmar2018image} introduces an image transformer model that adds the self-attention to an automatic regression model for image generation.
~\cite{wang2018non} formulates the self-attention as a non-local operation to model the spatial-temporal dependencies in video sequences.
~\cite{zhang2019self} proposes a self-attention GAN enforcing the generator to gradually consider non-local relationships in the feature space. It can learn to find long-range dependencies within internal representations of images.

Although our appearance transfer module (AT-module) is inspired by the self-attention, they are significantly different. The queries, keys and values in the AT-module are specially designed for pose-guided appearance transfer, and they are semantically different. By contrast, these items in the self-attention are obtained from the same input. As a result, the self-attention models the non-local relations within a single feature map while our AT-module finds the spatial correspondence between the  source stream and the target stream to perform appearance transfer.
\begin{figure*}[t]
\begin{center}
\includegraphics[width=\textwidth]{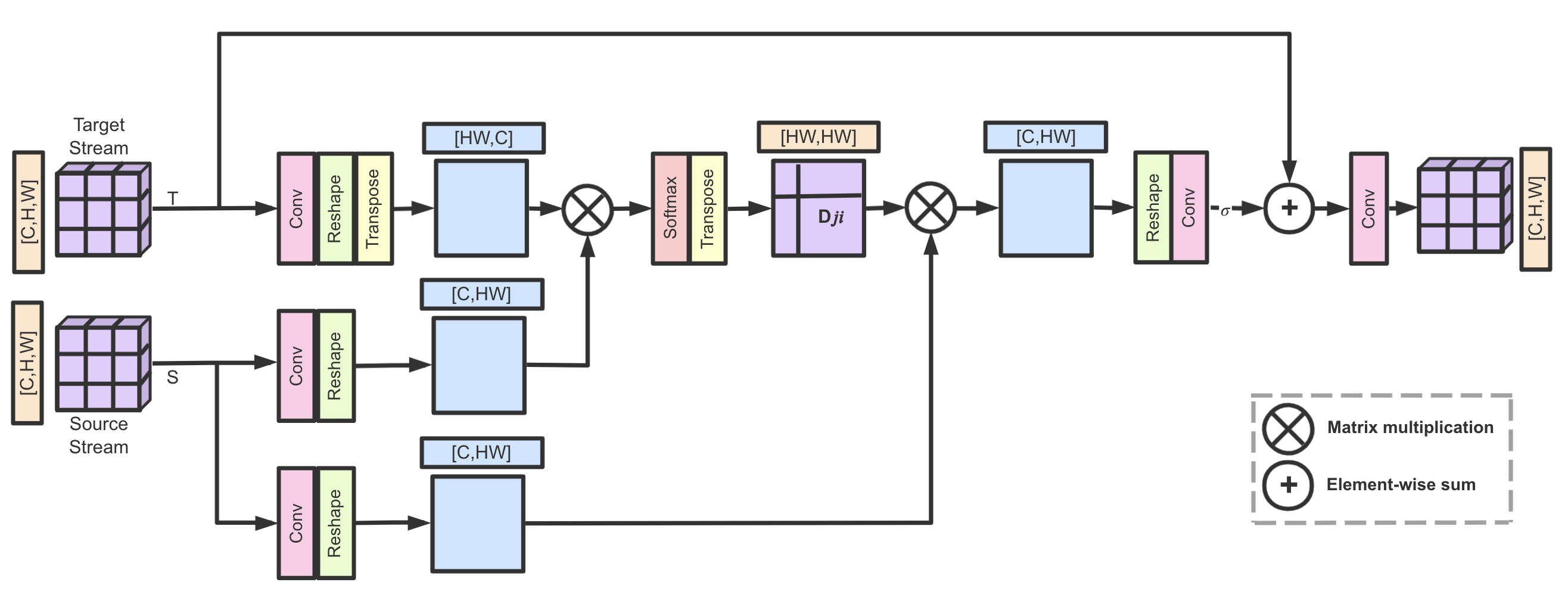}
\end{center}
\caption{Illustration of the proposed appearance transfer module (AT-module). It calculates the correspondence between the feature vector at each location in the target stream and the feature vector at each location in the source stream. Then the correspondence is used to transfer the appearance information from the source stream to the target stream. }
\label{AT-module}
\end{figure*}
%------------------------------------------------
\section{Our Approach}
\subsection{Overview of 2s-ATN}
As illustrated in Fig. \ref{generator}, our 2s-ATN is a multi-stage architecture consisting of two streams. The input of the target stream is the target pose $\mathbf{P}_t$. The input of the source stream is the concatenation of the source pose $\mathbf{P}_s$ and the source image $\mathbf{I}_s$. Both source and target poses are represented as keypoint heatmaps. The output of the network is a generated target image $\mathbf{I}_t$ containing the same person as the source image $\mathbf{I}_s$ but in the target pose $\mathbf{P}_t$.

The network first uses two encoders to produce initial feature maps for the two streams. Each encoder consists of two down-sampling convolutional layers, and they do not share weights. The initial source features contain both appearance and structure information while the initial target features contain only structure information.
Then, a cascade of AT-blocks progressively transfer the appearance from the source stream to the target stream guided by the structure information. Each block consists of an AT-module and several feature fusion modules. That is, they have the same architecture but do not share weights. Finally, the target feature map from the last AT-block passes through a decoder to generate the target image. The decoder consists of two deconvolutional layers. We will detail the AT-block in Sec. \ref{sec_block} and the loss function in Sec. \ref{sec_loss}.

\subsection{Appearance Transfer Block (AT-Block)}\label{sec_block}
As shown in Fig.~\ref{generator}, an AT-block takes as input the two-stream feature maps $\mathbf{F}_s\in \mathbb{R}^{C\times H\times W}$ and $\mathbf{F}_t\in \mathbb{R}^{C\times H\times W}$ obtained from the previous block or the encoder and outputs their updated feature maps $\mathbf{F}'_s\in \mathbb{R}^{C\times H\times W}$ and $\mathbf{F}'_t\in \mathbb{R}^{C\times H\times W}$. Here $C$, $H$ and $W$ respectively denote the channels, height and width of a feature map, and the subscripts $s$ and $t$ respectively indicate the source and target streams. 
An AT-block consists of an AT-module and several two-stream feature fusion modules, which are detailed below.

{\bf Appearance transfer module (AT-module).} The pipeline of an AT-module is illustrated in Fig. \ref{AT-module}. We first pass the two-stream feature maps $\mathbf{F}_s$ and $\mathbf{F}_t$ through convolutions and reshape the results as $\mathbf{S}\in \mathbb{R}^{C\times HW}$ and $\mathbf{T}\in \mathbb{R}^{C\times HW}$, respectively. Then we feed them into $1\times 1$ convolution layers (implemented as matrix multiplications) to produce three matrices $\mathbf{K}\in \mathbb{R}^{\bar{C}\times HW}$, $\mathbf{V}\in \mathbb{R}^{\hat{C}\times HW}$ and $\mathbf{Q}\in \mathbb{R}^{\bar{C}\times HW}$:
\begin{align}
\mathbf{K}&=\mathbf{W}_k\mathbf{S} \\
\mathbf{V}&=\mathbf{W}_v\mathbf{S} \\
\mathbf{Q}&=\mathbf{W}_q\mathbf{T}
\end{align}
where $\mathbf{W}_k$,${\mathbf{W}_q}\in \mathbb{R}^{\bar{C}\times C}$, $\mathbf{W}_v\in \mathbb{R}^{{\hat{C}}\times C}$ are learnable weight matrices. We set $\bar{C}=C/8$, ${\hat{C}}=C/2$ for memory efficiency, and it does not cause a significant performance drop.
Each column of $\mathbf{K}$, $\mathbf{V}$ or $\mathbf{Q}$ is a \emph{key}, a \emph{value} or a \textit{query} respectively. Our AT-module means to match (target) queries  to the (source) keys  and then use the correspondence to transfer the relevant (source) values from the source stream to the target stream.

To achieve this goal, we first obtain a correspondence map $\mathbf{D}\in \mathbb{R}^{HW\times HW}$ by applying a softmax normalization to each row of $\mathbf{Q}^T\mathbf{K}$:
\begin{equation}
\mathbf{D}_{ij}=\frac{\exp({\mathbf{Q}_i^T} {\mathbf{K}}_j)}{\sum_{j=1}^{HW}{\exp({{\mathbf{Q}}_i^T}{{\mathbf{K}}_j})}} 
\end{equation}
where $\mathbf{D}_{ij}$ is the $(i,j)$th element of $\mathbf{D}$, $\mathbf{Q}_i$ is the $i$th column of $\mathbf{Q}$, $\mathbf{K}_j$ is the $j$th column of $\mathbf{K}$. 
$\mathbf{D}_{ij}$ is a \emph{soft} correspondence score between the $i$th query, i.e., the $i$th position in the source feature map, and the $j$th key, i.e., the $j$th position in the target feature map. We can interpret the $i$th row of $\mathbf{D}$ as a probability distribution of each key matching the $i$th query. The correspondence map serves as the basis of appearance transfer.

Then we retrieve the value for the $i$th query as a linear combination of the columns of $\mathbf{V}$ weighted by the $i$th row of $\mathbf{D}$. A matrix $\mathbf{W}_o\in\mathbb{R}^{C\times \hat{C}}$ is multiplied to the retrieved values to increase their dimension:
\begin{equation}
\mathbf{A}= \mathbf{W}_o\mathbf{V}\mathbf{D}^T
\end{equation}
where $\mathbf{A}\in \mathbb{R}^{C\times HW}$ is the appearance information to be transferred from the source stream to the target stream. During the \emph{query-and-transfer} process, the source appearance is aligned with the target pose. Since the alignment is non-local, our AT-module can handle large pose transform.

Not all content of the target image can be found in the source image because of occlusion. To encourage the target stream to generate new content that can not be found in the source stream, we update  $\mathbf{A}$  by multiplying it with a scale parameter and adding back the target stream $\mathbf{T}$:
\begin{equation}
\mathbf{A}'=\sigma\mathbf{A}+\mathbf{T}
\end{equation}
where $\sigma$ is a learnable scalar.

\textbf{Two-stream feature fusion modules.} As shown in Fig. \ref{generator}, the features in the target stream are updated by fusing the features in the source stream and the transferred appearance. We also add a residual connection \cite{he2016deep} to ease the training process. 
The features in the source stream are updated by fusing the new target features. Two common choices of the fusion function are summation and concatenation. Our ablation study indicates the latter works better for both fusion modules. The two-stream feature fusion modules are important as they allow local information exchange between the two streams, which supplements the non-local appearance transfer. 
\begin{figure*}[t]
	\begin{center}
		\includegraphics[width=\textwidth]{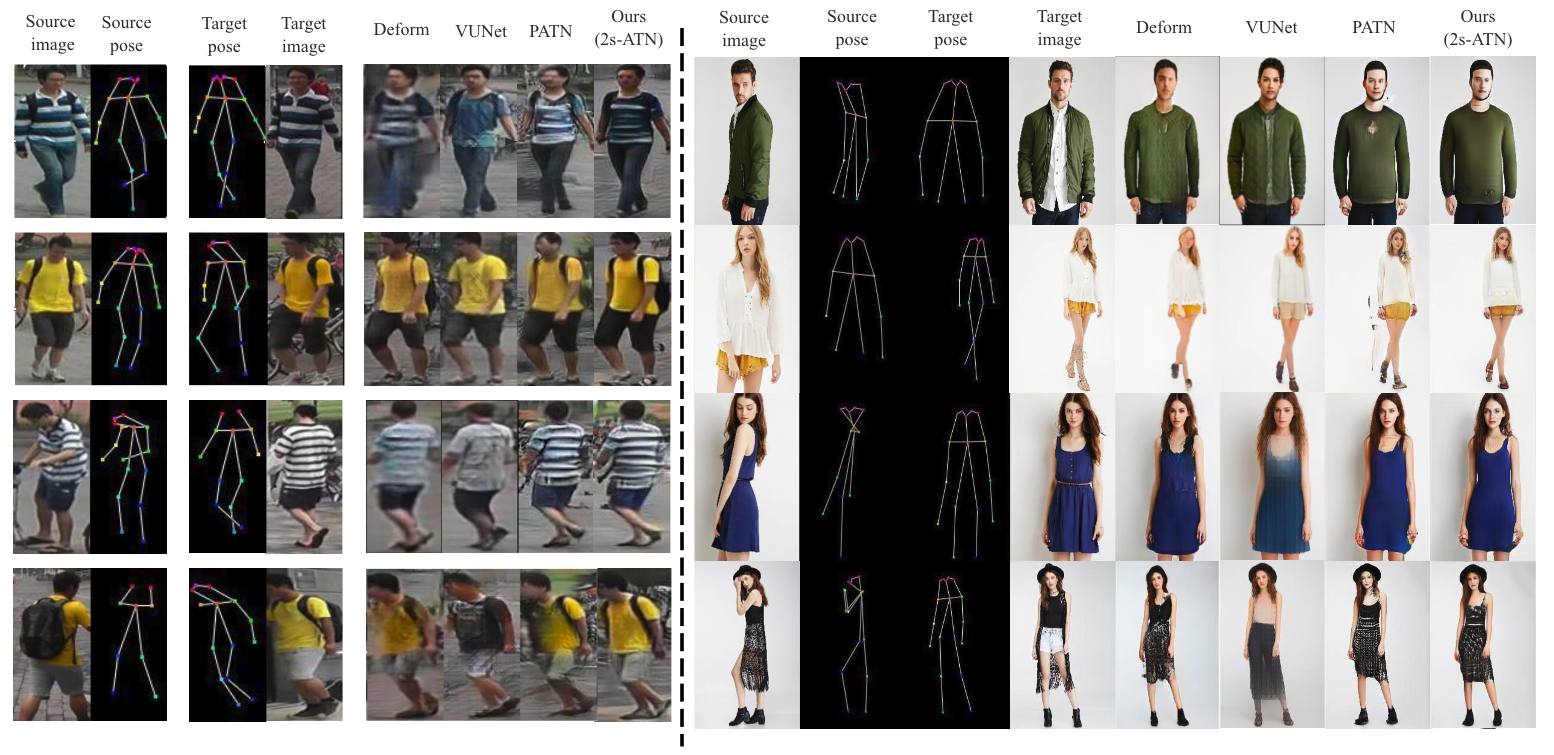}
	\end{center}
	\caption{Qualitative comparison on Market-1501 and DeepFashion.}
	\label{figure_results}
\end{figure*}
\subsection{Loss Function}\label{sec_loss}
The full loss function is:
\begin{equation}
\mathcal{L}={ \arg\min_G\max_D} ~{\alpha_g\mathcal{L}_{GAN}+{\alpha_1\mathcal{L}}_1+{\alpha_p\mathcal{L}}_p}\\
\end{equation}
where $\mathcal{L}_{GAN}$, $\mathcal{L}_1$ and $\mathcal{L}_p$ respectively denote the adversarial loss, the ${\ell}_{1}$-norm loss and the perceptual loss, and $\alpha_g$, $\alpha_1$ and $\alpha_p$ represent their respective weights.
$\mathcal{L}_1$ calculates the $\ell_1$-norm distance between the generated image $\mathbf{I}_t$ and the ground truth target image $\mathbf{I}_{gt}$: $\ell_1=||\mathbf{I}_{gt}-\mathbf{I}_t||_1$.
The perceptual loss $\mathcal{L}_p$ has been widely used for image generation and translation \cite{esser2018variational,siarohin2018deformable,ledig2017photo,johnson2016perceptual} as it helps generate more realistic and smoother images.
It is defined as:
\begin{equation}
\mathcal{L}_{p}=\frac{1}{W_\rho H_\rho C_\rho}||\phi_\rho(\mathbf{I}_{gt})- \phi_\rho(\mathbf{I}_t)||_1
\end{equation}
where $\phi_\rho$ is the output of the conv1\_2 layer from the VGG-19 model~\cite{simonyan2014very} pretrained on ImageNet~\cite{russakovsky2015imagenet}, and $W_\rho, H_\rho, C_\rho$ are the width, height and depth of $\phi_\rho$, respectively.
We adopt the adversarial loss introduced in \cite{zhu2019progressive}. 
It consists of an appearance discriminator and a shape discriminator to determine the possibility that the generated image contains the same person in the input image and the degree to which the generated image is aligned with the target pose.
%It is defined as
\begin{table*}[t]%开始一个表格environment
\begin{center}
%\centering
\resizebox*{\textwidth}{!}{
        \begin{tabular}{c|c|c|c|c|c|c|c|c}
        \hline  
        \multirow{2}{*}{Method} &\multicolumn{5}{c|}{Market-1501}&\multicolumn{3}{c}{DeepFashion} \\ \cline{2-9}
        & SSIM & IS & Mask-SSIM & Mask-IS & PCKh & SSIM & IS & PCKh\\
        \hline 
        DPIG ~\cite{ma2018disentangled} & 0.099 &  3.483 & 0.614 & 3.491 & $-$ & 0.614 & 3.228 & $-$\\
        %\cdashline{}
        VUNet ~\cite{esser2018variational} & 0.266 & 2.965 & 0.793 & 3.549 & 0.92 & 0.763 & $\mathbf{3.440}$ & 0.93\\
        %\cdashline{}
        Deform~\cite{siarohin2018deformable} & 0.290 & 3.185 & 0.805 & 3.502 & $-$ & 0.756 & 3.439 & $-$\\
        %\cdashline{}
        PATN~\cite{zhu2019progressive} & 0.311 & 3.323 & 0.811 & 3.773 & 0.94 & 0.773 & 3.209 & 0.96\\
        %\cdashline{}
        BFT~\cite{albahar2019guided} & $-$ & $-$ & $-$ & $-$ & $-$ & 0.767 & 3.220 & $-$\\
        %\cdashline{}
        C2GAN~\cite{tang2019cycle} & 0.282 & 3.349 & 0.811 & 3.510 & $-$ & $-$ & $-$ & $-$\\
        %\cdashline{}
        ADG~\cite{men2020controllable} & $-$ & $-$ & $-$ & $-$ & $-$ & 0.772 & 3.364 & $-$\\
        %\cdashline{}
        APS~\cite{huang2020generating} & 0.312 & 3.132 & 0.808 & 3.729 & 0.94 & 0.775 & 3.295 & 0.96\\
        \hline 
        PATN* ~\cite{zhu2019progressive} & 0.301 & 3.344 & 0.805 & 3.773 & 0.94 & 0.767 & 3.209 & 0.96\\
        %\cdashline{}
            {\bf Ours}  & $\mathbf{0.320}$ & $\mathbf{3.504}$ & $\mathbf{0.813}$ & $\mathbf{3.845}$ & $\mathbf{0.94}$ & $\mathbf{0.775}$ & 3.206 & $\mathbf{0.96}$\\
        %\cdashline{}
           Real Data  & 1.000 & 3.890 & 1.000 & 3.706 & 1.00 & 1.000 & 4.053 & 1.00\\
        \hline  
\end{tabular}}
\end{center}
\caption{Quantitative results on Market-1501 and DeepFashion. ($*$) denotes the results reproduced by us. For all metrics, higher values indicate better performance.}
\label{banchmark}
\end{table*}

\begin{table*}%开始一个表格environment
\begin{center}
%\resizebox{0.47\textwidth}{!}{
        \begin{tabular}{c|c|c}%设置了每一列的宽度，强制转换。  
        \hline  
        \multirow{1}{*}{Method} & \#Parameters & Speed\\ %用&来分隔单元格的内容 \\表示进入下一行 
        \hline 
    PG2 ~\cite{ma2017pose}& 437.09 M & 10.36 fps\\
    Deform ~\cite{siarohin2018deformable}& 82.08 M & 17.74 fps\\
    VUNet ~\cite{esser2018variational} & 139.36 M & 29.37 fps\\
    PATN ~\cite{zhu2019progressive}&  \textbf{41.36 M} & 60.61 fps\\
    \textbf{Ours} &43.28 M & $\mathbf{74.21 fps}$\\
    \hline  
\end{tabular}
%}
\end{center}
\caption{Comparison of model sizes and inference speeds on DeepFashion.}
\label{Speed}
\end{table*}

\section{Experiment}
In this section, we conduct extensive experiments both qualitatively and quantitatively. They will demonstrate that our 2s-ATN outperforms state-of-the-art methods regarding visual fidelity and alignment with targeted person poses.

{\bf Datasets.} We use two challenging person image datasets: Market-1501~\cite{zheng2015scalable} and DeepFashion~\cite{liu2016deepfashion}. The resolution of images in DeepFashion is higher ($256\times256$) than images in Market-1501 ($128\times64$).
We employ OpenPose~\cite{cao2017realtime} to detect human body joints. Both the source and target poses consist of an 18-channel heatmap encoding the positions of 18 human body joints . We have 263,632 pairs of training images from Market-1501, and 101,966 pairs from DeepFashion. Their testing sets contain 12,000 pairs and 8,570 pairs, respectively. Note the person identities of the training set do not overlap with those of the testing set.

\begin{figure*}[t]
	\begin{center}
		\includegraphics[width=1\textwidth]{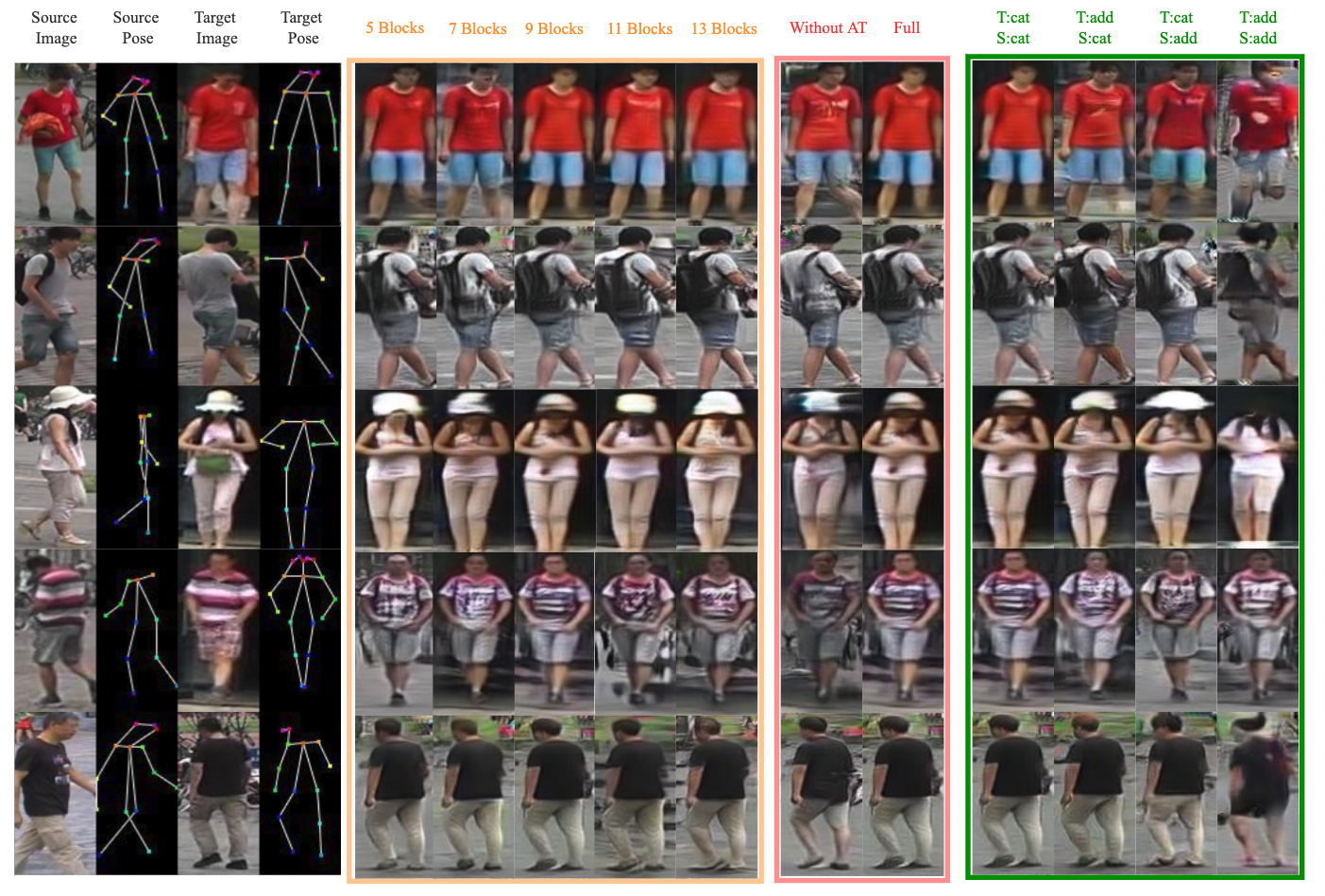}
	\end{center}
	\caption{ Ablation study of the proposed 2s-ATN on Market-1501. First column: input and ground truth output. Second column (orange): ablation study on the number of AT-blocks. Third column (red): ablation study on the AT-module. Fourth column (green): ablation study on the two-stream feature fusion modules.}
	\label{ablition}
\end{figure*}

%-------------------------------------------------------------------------
{\bf Evaluation metrics.} We follow~\cite{ma2017pose,siarohin2018deformable,zhu2019progressive} and adopt Structure Similarity (SSIM)~\cite{wang2004image}, Inception Score (IS)~\cite{salimans2016improved}, and their masked versions, i.e., Mask-SSIM and Mask-IS, as the evaluation metrics. Moreover, we adopt the PCKh score proposed in~\cite{zhu2019progressive} to assess the shape consistency explicitly.

{\bf Implement details.} Our method is implemented in PyTorch using two NVIDIA GeForce RTX 2080 Ti GPUs. The Adam optimizer~\cite{kingma2014adam} is adopted to train the proposed model for around 90k iterations with $\beta_1=0.5$, $\beta_2=0.999$. The learning rate is fixed as 0.0001 in the first 60k iterations and then linearly decayed to 0 in the last 30k iterations. We use 9 AT-blocks in the generator for both datasets. For the hyper-parameters, ($\alpha_g$,$\alpha_{1}$,$\alpha_{p}$) are set as (20, 17, 17) for DeepFashion and (20, 17, 17) for Market-1501,  respectively. Instance normalization~\cite{ulyanov2016instance} is applied for both datasets. The batch size is set as 7 for DeepFashion and 32 for Market-1501. 
Dropout~\cite{hinton2012improving} is only used in the AT-blocks, and the dropout rate is set to 0.5. Leaky ReLU~\cite{maas2013rectifier} is applied after every convolution or normalization layer in the discriminators, and its negative slope coefficient is set to 0.2. Each training epoch takes about 95 seconds for Market-1501 and 460 seconds for Deepfashion. In total, it takes about one day to train the network on Market-1501  and four days on DeepFashion.

%-------------------------------------------------------------------------

\subsection{Comparison with States-of-the-art Methods}
{\bf Quantitative and qualitative comparison.} We compare the proposed 2s-ATN with several state-of-the-art methods such as DPIG ~\cite{ma2018disentangled}, VUnet~\cite{esser2018variational}, Deform~\cite{siarohin2018deformable}, PATN~\cite{zhu2019progressive}, BTF~\cite{albahar2019guided}, C2GAN~\cite{tang2019cycle}, ADG~\cite{men2020controllable} and APS~\cite{huang2020generating}. Table~\ref{banchmark} shows the quantitative results measured by SSIM, IS, Mask-SSIM, Mask-IS, and PCKh metrics. 
The 2s-ATN achieves the best performance under most metrics on the two datasets. Fig. \ref{figure_results} shows that the appearance or texture generated by the proposed method is more consistent and appealing than the others.

We also take PATN ~\cite{zhu2019progressive} as a strong baseline and reproduce its results via the code provided by the authors.
Compared with PATN, 2s-ATN has achieved better performance under the SSIM, IS, Mask-SSIM, and Mask-IS metrics and the same performance under PCKh. 
However, we observe that compared with the images generated by PATN~\cite{zhu2019progressive}, the images generated by 2s-ATN are more realistic and have fewer visual artifacts (as Fig.~\ref{figure_results} depicted).

{\bf Comparison of model size and inference speed.} Tab.~\ref{Speed} compares the model size and (single-GPU) inference speed of our 2s-ATN with those of four state-of-the-art methods. The GPU time is reported. Thanks to the simple and clean structure of our 2s-ATN, it achieves the highest processing speed with a much smaller or comparable model size.

\subsection{Ablation Study}
In this section, we perform ablation study to analyze the impact of each component in our model on the performance. 
%We first describe the variants obtained by incrementally removing components from the full framework (as depicted in Fig.~\ref{ablition}). 
There are two kinds of essential modules in the 2s-ATN: the AT-module and the two-stream feature fusion modules.

\begin{table}%开始一个表格environment
\begin{center}
        \centering
        \resizebox{0.47\textwidth}{!}{
        \begin{tabular}{c|c|c|c|c}%设置了每一列的宽度，强制转换。  
        \hline  
        \multirow{2}{*}{Method} & \multicolumn{4}{c}{Market-1501} \\ %用&来分隔单元格的内容 \\表示进入下一行  
        \cline{2-5}
        & SSIM & IS & Mask-SSIM & Mask-IS\\
        \hline{}
        2s-ATN (Full) & $\mathbf{0.320}$ & $\mathbf{3.504}$ & $\mathbf{0.813}$ & $\mathbf{3.845}$\\
        %\cdashline{}
        w/o AT-module & 0.313 & 3.252 & 0.812 & 3.793\\
        \hline  
\end{tabular}}
\end{center}
\caption{Ablation study on the AT-module.}
\label{without AT}
\end{table}

{\bf Effect of the AT-module.} The results of this ablation study are shown in Tab.~\ref{without AT} and Fig.~\ref{ablition}. Removing the AT-modules from our 2s-ATN will decrease SSIM, IS, Mask-SSIM and Mask-IS from 0.320, 3.504, 0.813, and 3.845 to  0.313, 3.252, 0.812, and 3.793, respectively. 
The qualitative results show the images generated by the full model look much better than those generated by the model without the AT-module.
Thus the proposed AT-module helps generate photo-realistic person images as it enables the network to perform non-local spatial manipulation.

\begin{table}%开始一个表格environment
\begin{center}
        \centering
        \resizebox{0.47\textwidth}{!}{
        \begin{tabular}{c|c|c|c|c}%设置了每一列的宽度，强制转换。  
        \hline  
        \multirow{2}{*}{Method} & \multicolumn{4}{c}{Market-1501} \\ 
        \cline{2-5}
        & SSIM & IS & Mask-SSIM & Mask-IS\\
        \hline{}
       T:cat, S:cat & $\mathbf{0.320}$ & 3.504 & $\mathbf{0.813}$ & $\mathbf{3.845}$\\
        %\cdashline{}
        T:add, S:cat & 0.320 & 3.335 & 0.813 &  3.815\\
        %\cdashline{}
        T:cat, S:add & 0.309 &3.460 & 0.809 &3.843\\
        %\cdashline{}
        T:add, S:add & 0.234 & $\mathbf{3.894}$ &0.768 & 3.675\\
        \hline  
\end{tabular}}  
\end{center}
\caption{Ablation study on the two-stream feature fusion modules. ``T" and ``S" denote the target stream and source stream, respectively. }
\label{adds/cats}
\end{table}
{\bf Effect of two-stream feature fusion modules.}  Qualitative and quantitative results are shown in Fig.~\ref{ablition} and Tab.~\ref{adds/cats}. 
``T" and ``S" denote the two fusion modules in the target stream and the source stream, respectively. ``Cat" and ``add" are short for concatenation and summation, respectively.
The quantitative results in Tab.~\ref{adds/cats} indicate concatenation generally works better than summation for both fusion modules. The visualization in Fig. ~\ref{ablition} shows using concatenation in both streams can help our 2s-ATN generate more appealing images  than other combinations of fusion methods. The shape and appearance of generated persons are more consistent with those in the ground truth. 

\begin{table}%开始一个表格environmen
\resizebox{0.47\textwidth}{!}{
        \begin{tabular}{c|c|c|c|c}%设置了每一列的宽度，强制转换。  
        \hline 
        \multirow{2}{*}{\#AT-blocks} & \multicolumn{4}{c}{Market-1501} \\ %用&来分隔单元格的内容  
        \cline{2-5}
    & SSIM & IS & Mask-SSIM & Mask-IS\\
    \hline 
    5 & 0.317 & 3.354 & 0.813 & 3.783\\
    %\cdashline{}
    7 & 0.319 & 3.490 & 0.812 & 3.799\\
    %\cdashline{}
    9 & $\mathbf{0.320}$ & $\mathbf{3.504}$ & $\mathbf{0.813}$ & 3.845\\
    %\cdashline{}
    11 & 0.314 & 3.462 & 0.810 & 3.844\\
    %\cdashline{}
    13 &0.316& 3.323 & 0.813 & $\mathbf{3.852}$\\
    \hline  
    \end{tabular}}
\caption{Ablation study on the number of AT-blocks.}
\label{block_variation}
\end{table}

{\bf Effect of the number of AT-blocks.} Quantitative and qualitative results are shown in Tab.~\ref{block_variation} and Fig.~\ref{ablition}. We observe that the proposed generator works best when it consists of 9 AT-blocks. Increasing or decreasing the number of AT-blocks may result in slightly poor quantitative and qualitative performance. Therefore, we have used 9 AT-blocks as the default setting in the other experiments. It is worth noting that using only 5 AT-blocks still results in acceptable, through not the best, results both qualitatively and quantitatively. This indicates that a light version of the proposed method is still applicable when the computation resource is limited.

\section{Conclusion}
This paper introduces a two-stream appearance transfer network (2s-ATN) for pose guided person image generation. It is a novel multi-stage architecture consisting of a source stream and a target stream. Its core is a novel appearance transfer module (AT-module) in each stage. It learns to find the structure correspondence between the two-stream feature maps and then transfer the appearance information from the source stream to the target stream. We also supplement the non-local appearance transfer with the local two-stream feature fusion modules. Experimental results indicate that the proposed 2s-ATN can effectively handle large pose deformation and occlusion while retaining the texture details. 

\newpage

\bibliography{egbib}
\end{document}